\documentclass[letterpaper]{article} 
\usepackage{aaai2027}
\nocopyright

\usepackage[hyphens]{url}  
\usepackage{graphicx} 
\usepackage{natbib}  
\usepackage{caption} 
\usepackage{algorithm}
\usepackage{algorithmic}
\usepackage{amsmath}
\usepackage{amssymb}
\usepackage{multirow}
\usepackage{xspace}

\usepackage{newfloat}
\usepackage{listings}
\DeclareCaptionStyle{ruled}{labelfont=normalfont,labelsep=colon,strut=off} 
\floatstyle{ruled}
\newfloat{listing}{tb}{lst}{}
\floatname{listing}{Listing}

\usepackage{booktabs}

\title{INCLAIR: Inception-Based Longitudinal Clinical Anomaly Detection with Informed Reasoning}
\author {
    Maxx Richard Rahman\textsuperscript{\rm 1,\rm 2}\corresponding,
    Wolfgang Maass\textsuperscript{\rm 1,\rm 2}
}
\affiliations {
    \textsuperscript{\rm 1}German Research Center for Artificial Intelligence (DFKI), Germany\\
    \textsuperscript{\rm 2}Saarland University, Germany\\
    maxx\_richard.rahman@dfki.de, 
    wolfgang.maass@dfki.de
}

\begin{document}

\maketitle

\begin{abstract}
Detecting anomalies in longitudinal clinical profiles is clinically important but difficult: abnormal evidence is often sparse, patient histories have unequal length, and expert explanations are costly. We propose INCLAIR, a framework that scores each observation against multiple historical contexts, aggregates evidence at the profile level, and generates grounded natural-language explanations under limited expert supervision. Under stated within-profile exchangeability assumptions, the complete mean subsequence score takes an order-$l$ U-statistic form, yielding a variance decomposition and an incomplete-subset approximation that controls combinatorial inference cost independently of profile length. The same analysis shows that mean aggregation attenuates localized anomalies by a factor set by the anomaly support and profile length, motivating validation-selected top-$k$ pooling. Across three clinical datasets, INCLAIR consistently outperforms state-of-the-art baselines. We further validate practical relevance through a case study on longitudinal steroid profiles, comparing INCLAIR's predictions and explanations against domain-expert assessments supported by DNA analysis. The results show that INCLAIR enables clinically actionable anomaly detection under limited expert supervision.
\end{abstract}


\section{Introduction}
Longitudinal clinical profiles are sequences of biomarker measurements, physiological signals, or imaging-derived features collected over time, and are fundamental to clinical decision-making in disease-progression monitoring and treatment-response assessment~\cite{AlEwaidat2025PrecisionMedicine,Gkintoni2025AIAutismReview}. Representative settings include neurodegenerative monitoring from longitudinal neuroimaging~\cite{Sintini2020ADPhenotypes}, continuous physiological surveillance in intensive care~\cite{Stellpflug2021PhysMonitoring}, and steroid profiling in anti-doping and endocrine assessment~\cite{Equey2023ABPSteroids}. In these high-stakes settings, deviations from an individual's historical trajectory may reflect abnormal physiological responses or procedural errors and should be detected reliably, which strong temporal dependencies, high inter-subject variability, and scarce labelled anomalies make difficult.

Existing methods identify statistical irregularities or reconstruction errors using recurrent models~\cite{Cascarano2023LongitudinalMLReview}, autoencoders~\cite{Nawaz2024EnsembleAutoencoders}, or generative approaches~\cite{Loni2025GenerativeMedicalAI}. Ensemble detectors~\cite{Zhao2019LSCP,Zhao2021SUOD} improve robustness but remain agnostic to temporal structure and physiological context, and most operate at a single temporal scale, missing the short-term fluctuations and long-term trends that clinical interpretation requires~\cite{Lai2018LSTNet}. High sensitivity is therefore often obtained at the cost of excessive false positives, reducing clinical utility and increasing the burden of confirmatory testing.

Clinical practice also requires understanding \emph{why} a profile is anomalous, to guide intervention or confirmatory testing. Traditional time-series explainability such as attention or feature attribution~\cite{Choi2016RETAIN,lipton2018} offers limited insight into longitudinal physiological reasoning. Large language models (LLMs) can produce natural-language clinical explanations~\cite{Wang2023ClinicalGPT,Singhal2025ExpertMedicalQA}, but they lack grounding in structured temporal representations and require large volumes of expert-annotated rationales that are rarely available at scale, so under this scarcity they frequently produce clinically unfaithful reasoning. The gap is acute in high-stakes screening: confirmatory DNA analysis is highly certain for detecting sample swapping but too costly to apply to all suspicious cases~\cite{Bancos2020UrineSteroidMetabolomics}, invasive follow-up in neurodegenerative monitoring is reserved for high-risk patients, and alarm fatigue persists in critical care~\cite{Chan2025NeurodegenerativeDiseases}. These settings jointly demand accurate detection and clinically grounded reasoning under limited expert supervision. Key contributions:

\begin{itemize}
    \item We propose combinatorial history conditioning for longitudinal anomaly detection and show that the complete mean profile score is an order-$l$ U-statistic under explicit within-profile assumptions. 
    \item We develop a limited-supervision reasoning pipeline that selects a base language model, filters synthetic explanations by structured claim audits, and fine-tunes with LoRA on a small set of expert rationales.
    \item We show the practical applicability through experiments and a DNA-validated case study, highlighting its potential to reduce reliance on costly confirmatory tests.
\end{itemize}

\section{Related Works}
\subsection{Anomaly Detection in Clinical Profiles}
Recurrent and sequence models detect anomalies via forecast residuals or likelihood surrogates~\cite{Malhotra2016LSTMPrognostics,Hundman2018SpacecraftAnomalies}, while autoencoder variants such as USAD optimize coupled reconstruction losses to separate normal from anomalous patterns~\cite{Audibert2020USAD}. Probabilistic and latent-variable methods score anomalies using reconstruction error or approximate likelihood, e.g., Beta-VAE~\cite{Higgins2017BetaVAE}, and GAN-based detectors such as AnoGAN use latent search and reconstruction consistency~\cite{Schlegl2017GANAnomalyDetection}. Classical baselines remain competitive in tabular or low-dimensional settings, e.g., IsoForest~\cite{Liu2012IsolationForest}, and ensemble frameworks such as LSCP~\cite{Zhao2019LSCP} and SUOD~\cite{Zhao2021SUOD} aggregate heterogeneous detectors to stabilize unsupervised performance. SACNN~\cite{Rahman2024SACNN} emphasizes temporal locality and multi-scale pattern extraction for clinical time-series classification.

\subsection{Informed Clinical Reasoning}
Clinical interpretability for sequence models has relied on attention-based rationales (RETAIN~\cite{Choi2016RETAIN}) and post-hoc feature importance, though such explanations are often unfaithful to model decision logic~\cite{lipton2018}. LLMs have since enabled natural-language clinical reasoning through instruction tuning and domain adaptation, including specialized and open biomedical models (Med-PaLM~\cite{Singhal2025ExpertMedicalQA}, ClinicalGPT~\cite{Wang2023ClinicalGPT}, Meditron~\cite{Chen2023Meditron70B}, MedAlpaca~\cite{Han2023MedAlpaca}, AlpaCare~\cite{Zhang2023AlpacaRE}), while strong general LLMs (Mistral-7B~\cite{Jiang2024Mixtral}, Qwen2.5~\cite{Hui2024Qwen25Coder}, Phi~\cite{Abdin2024Phi4}, Llama 3.1~\cite{Grattafiori2024LLaMA3}) are often competitive with careful prompting. However, clinical evaluations indicate that factual grounding and faithfulness remain unreliable when task-specific expert supervision is scarce~\cite{Singhal2025ExpertMedicalQA,Wang2023ClinicalGPT,Zheng2023MTBench}.

\section{Problem Formulation}
\subsection{Anomaly Detection in Longitudinal Clinical Profiles}
Let $\mathcal{D} = \{\mathbf{X}_1, \dots, \mathbf{X}_N\}$ be a collection of longitudinal profiles, where each profile $\mathbf{X}_i = [\mathbf{x}_{i1}, \dots, \mathbf{x}_{iT_i}]$ consists of $T_i$ observations and $\mathbf{x}_{ij} \in \mathbb{R}^p$ is the vector of $p$ clinical variables. The entry $x_{ij\ell}$ is the value of the $\ell^{th}$ variable for individual $i$ at time point $j$, with $j = 1, \dots, T_i$ and $\ell = 1, \dots, p$. The detection task is to infer, for each profile $\mathbf{X}_i$, a binary indicator $y_i \in \{0,1\}$, where $y_i = 1$ denotes clinically meaningful abnormal behavior.

\subsection{Informed Reasoning under Limited Expert Supervision}
Let $\mathbf{e}_i$ denote a natural-language explanation for profile $\mathbf{X}_i$, describing the clinical or physiological rationale underlying $y_i$. Expert reports are available for only a small subset of profiles $\mathcal{D}_{gold} = \{ (\mathbf{X}_i, y_i, \mathbf{e}_i) \}_{i=1}^{N_g}$, with $N_g \ll N$, while the remaining profiles in $\mathcal{D}_{probe}$ are unlabeled and lack explanatory annotations. The problem is to learn a model that generalizes from this limited labeled and explained set to unseen profiles, producing explanations aligned with expert reasoning despite the scarcity of annotations.

\begin{figure*}[hbt!]
\centering{\includegraphics[width=0.9\linewidth]{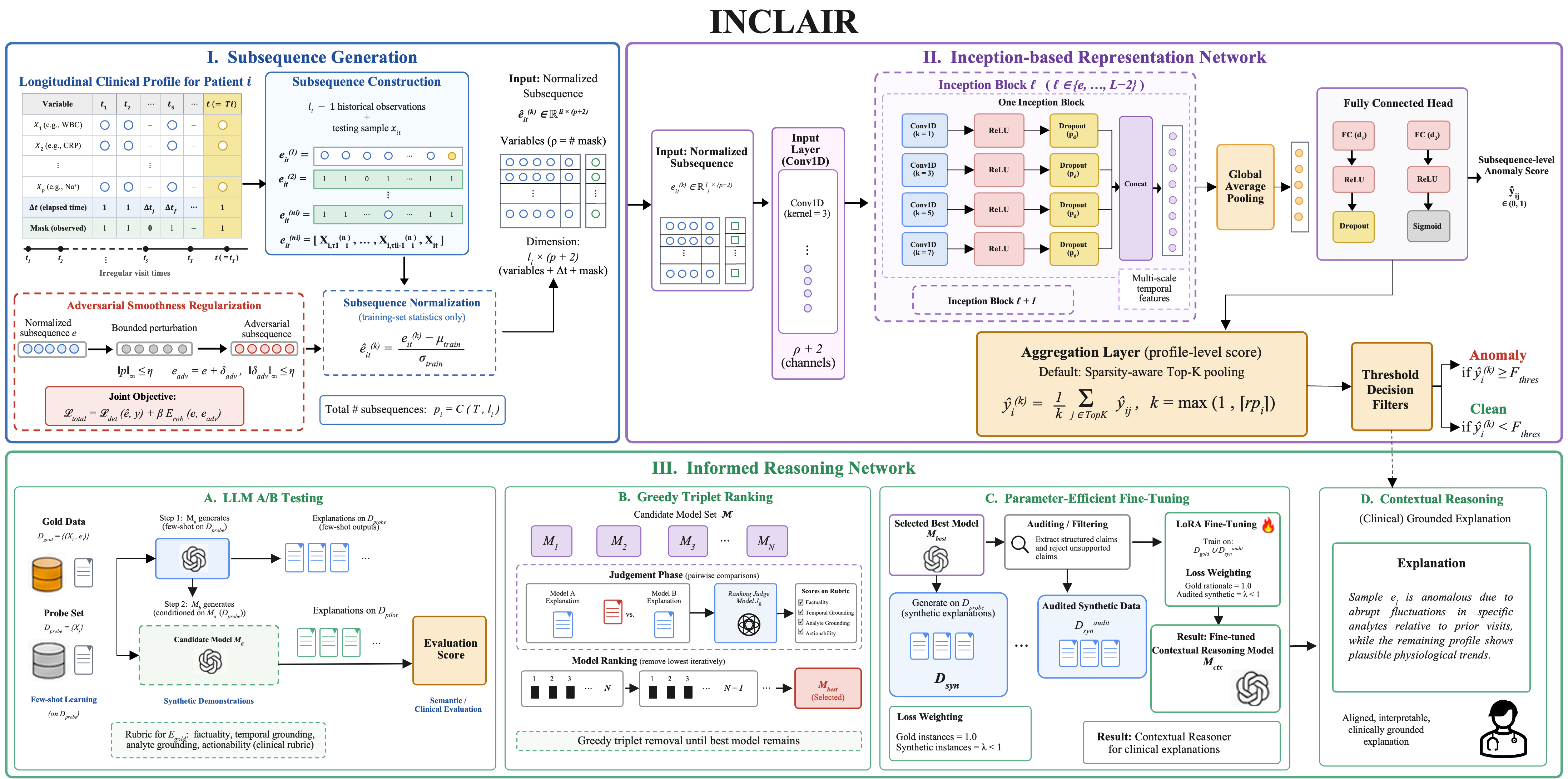}}
\caption{INCLAIR: i) Subsequence generator, ii) Inception-based representation network, iii) Informed reasoning network.
\label{inclair}}
\end{figure*}

\section{Methodology}
\subsection{Subsequence Generation}
Given a profile, we construct subsequences centered on a testing sample while preserving temporal ordering (Fig.~\ref{inclair}). For a fixed effective length $l_i$, each subsequence selects $l_i-1$ historical observations preceding a testing time point $t$, with the testing sample $\mathbf{x}_{it}$ as the final element. The generator maps $\mathbf{X}_i$ to $\mathcal{E}(\mathbf{X}_i) = \bigcup_{t=l_i}^{T_i} \mathcal{E}_t(\mathbf{X}_i)$, where each subsequence $\mathbf{e}_{i,t}^{(k)} \in \mathbb{R}^{l \times (p+2)}$ includes the clinical variables, $\Delta t$, and the mask channel:
$\mathbf{e}_{i,t}^{(k)} =
\big[\mathbf{x}_{i\tau_1^{(k)}}, \dots,
\mathbf{x}_{i\tau_{l_i-1}^{(k)}}, \mathbf{x}_{it} \big]$,
with $\tau_1^{(k)} < \dots < \tau_{l_i-1}^{(k)} < t$ and $\{\tau_1^{(k)}, \dots, \tau_{l_i-1}^{(k)}\} \subset \{1, \dots, t-1\}$. For each $t$, the generator enumerates all combinations of $l_i-1$ preceding observations, fixing the subsequence length after padding, preserving temporal behaviour, and letting the model compare the testing sample against multiple historical contexts. The total number of subsequences for profile $\mathbf{X}_i$ is
\begin{equation}
p_i = \sum_{t=l_i}^{T_i} \binom{t-1}{l_i-1}=\binom{T_i}{l_i},
\end{equation}
by the hockey-stick identity. We use a single length $l$ for all profiles, selected on validation. Since the datasets have profile length at most 20 (Table~\ref{tab:datasets}), the worst case is $\binom{20}{5}=15{,}504$ subsequences per profile and the per-profile average is far smaller, so full combinatorial enumeration is tractable and used as the default. Where $\binom{T_i}{l}$ is prohibitively large, we instead draw $B$ subsets uniformly at random without replacement per profile; Corollary~1 shows that $B=O(\epsilon^{-2}\log(1/\delta))$ samples approximate the complete-mean score to accuracy $\epsilon$ independently of $\binom{T_i}{l}$, bounding memory and inference cost. Each subsequence is normalized using training-set statistics and expressed relative to the patient's historical context.

\subsection{Inception-Based Representation Network}
The inception network learns discriminative representations from fixed-length subsequences by capturing temporal patterns at multiple resolutions. Given a normalized subsequence $\mathbf{e}_{i,j} \in \mathbb{R}^{l \times (p+2)}$, it produces a latent representation reflecting both short-term fluctuations and longer-term dependencies across clinical variables.

\textbf{Input Layer.}
Each subsequence $\mathbf{e}_{i,j}$ is passed through an input convolutional layer that projects the raw measurements into a higher-dimensional feature space via a one-dimensional convolution along the temporal axis, $\mathbf{h}_{i,j}^{(0)} = \sigma(\mathbf{e}_{i,j} * \mathbf{W}^{(0)} + \mathbf{b}^{(0)})$, where $*$ is the convolution operator, $\mathbf{W}^{(0)}, \mathbf{b}^{(0)}$ are learnable parameters, and $\sigma(\cdot)$ is a nonlinear activation. The input is the full subsequence tensor of shape $l \times (p+2)$, comprising the $p$ clinical variables, the elapsed-time channel $\Delta t$, and the padding-mask channel; the convolution maps it to a feature map with 16 channels.

\textbf{Inception Block.}
To capture multi-scale temporal patterns, INCLAIR uses inception blocks of parallel convolutional branches with different kernel sizes. Given a feature map $\mathbf{h}$, each branch applies a convolution with kernel size $k_q$,
$\mathbf{h}^{(q)} = \sigma(\mathbf{h} * \mathbf{W}^{(q)} + \mathbf{b}^{(q)}), \quad q = 1, \dots, Q$,
where $Q$ is the number of branches, and the branch outputs are concatenated along the feature dimension,
$\mathbf{h}^{\text{inc}} = \text{Concat}(\mathbf{h}^{(1)}, \mathbf{h}^{(2)}, \dots, \mathbf{h}^{(Q)})$.
This jointly models short- and long-range temporal dependencies without significantly increasing computational cost, and blocks can be stacked to progressively refine representations.

\textbf{Fully Connected Layer.}
The final inception block output is reduced by global average pooling followed by a fully connected layer. Let $\mathbf{z}_{i,j}$ be the pooled representation of $\mathbf{e}_{i,j}$. The subsequence-level anomaly score is $\hat{y}_{i,j} = \sigma(\mathbf{W}_f \mathbf{z}_{i,j} + b_f)$, where $\mathbf{W}_f, b_f$ are learnable and $\hat{y}_{i,j} \in [0,1]$ is the predicted likelihood that the subsequence is anomalous.

\textbf{Aggregation Layer.}
Because predictions are obtained at the subsequence level, an aggregation layer infers a profile-level decision. Mean aggregation suits diffuse profile shifts but dilutes localized anomalies, so we adopt sparsity-aware top-$k$ pooling as the default:
$ 
\hat{y}_i^{(k)} = \frac{1}{k_i} \sum_{j \in \operatorname{TopK}(\{\hat{y}_{i,1},\dots,\hat{y}_{i,p_i}\}, k_i)} \hat{y}_{i,j},
$
where $k_i=\max(1,\lceil r p_i\rceil)$ and $r$ is validation-selected. We report mean, max, and top-$k$ pooling in the ablation. A profile is classified anomalous if $\hat{y}_i^{(k)} \geq F_{\text{thres}}$, where $F_{\text{thres}}$ is chosen on validation to meet a prespecified specificity target.

\textbf{U-statistic Profile Scoring.}
The subsequence construction induces a U-statistic structure for the complete mean score. Let $S\subseteq[T_i]$ be a time-ordered index subset with $|S|=l_i$, and let $f_\theta(\mathbf{x}_S)$ be the subsequence score. The complete mean profile score is
$
\bar{y}_i=\binom{T_i}{l_i}^{-1}\sum_{S\subseteq[T_i], |S|=l_i} f_\theta(\mathbf{x}_S),
$
an order-$l_i$ U-statistic with a bounded, possibly asymmetric kernel. The following use standard U-statistic theory~\cite{Hoeffding1948UStatistics,Lee1990UStatistics} under these assumptions.

\noindent\textbf{Proposition 1. Variance decomposition.}
Let $\zeta_c=\operatorname{Var}(\mathbb{E}[f_\theta(\mathbf{x}_S)\mid c \text{ fixed coordinates}])$ for $c=1,\dots,l_i$. Then
\begin{equation}
\begin{aligned}
\operatorname{Var}(\bar{y}_i)
&=\binom{T_i}{l_i}^{-1}
\sum_{c=1}^{l_i}
\binom{l_i}{c}
\binom{T_i-l_i}{l_i-c}\zeta_c
=\frac{l_i^2}{T_i}\zeta_1+O(T_i^{-2}).
\end{aligned}
\end{equation}
Profile-score variance thus decreases with longer histories and increases quadratically with the effective window length, formalizing why very short profiles yield unstable decisions.

\noindent\textbf{Corollary 1. Incomplete U-statistic approximation.}
Let $\bar{Y}_{i,B}$ average $B$ subsets sampled uniformly with replacement from all $l_i$-subsets of a fixed observed profile. Conditional on $\mathbf{X}_i$, the sampled subset scores are independent draws from the finite empirical set of complete-subset scores, so by Hoeffding's inequality
$
\Pr(|\bar{Y}_{i,B}-\bar{y}_i|\geq\epsilon\mid \mathbf{X}_i)
\leq 2\exp(-2B\epsilon^2).
$
Hence $B=O(\epsilon^{-2}\log(1/\delta))$ sampled subsequences suffice for an $\epsilon$-accurate approximation to the complete mean score with probability at least $1-\delta$, independent of $\binom{T_i}{l_i}$. This is a Monte Carlo guarantee conditional on the observed profile; it does not remove biological dependence between visits or guarantee generalization across patients.

\noindent\textbf{Proposition 2. Sparse anomaly attenuation.}
Suppose $m$ of $T$ time points are anomalous, and the kernel has mean $\mu_a$ when a subset intersects the anomalous support and $\mu_n$ otherwise. Under mean pooling,
\begin{equation}
\mathbb{E}[\bar{y}_i]=\mu_n+(\mu_a-\mu_n)q,\quad
q=1-\binom{T-m}{l}/\binom{T}{l}.
\end{equation}
For a single anomalous observation, $q=l/T$. Mean pooling thus attenuates a localized signal by the fraction of windows containing the anomalous time point; top-$k$ pooling is designed to recover these high-scoring windows rather than average them away.

\noindent\textbf{Scope of Top-$k$ Pooling.}
Top-$k$ pooling is not itself a complete U-statistic, since it depends on the order statistics of the learned subset scores. We use it as a screening rule motivated by Proposition~2 and selected only on validation. When anomalous windows occupy fraction $q$ of all windows and exceed normal-window scores by a clear margin, a rate $r\leq q$ concentrates the pooled score on the anomalous support; if $r>q$ or the separation is weak, it imports normal windows and loses its advantage over the mean.

\textbf{Adversarial Smoothness Regularization.}
We use a universal bounded perturbation as a smoothness regularizer, not a proxy for natural distribution shift. Let $\mathbf{e} \in \mathbb{R}^{l \times (p+2)}$ be a normalized subsequence and $f(\mathbf{e})$ the model prediction. The perturbation $\boldsymbol{\rho}$ is updated by one projected gradient-ascent step on the detection loss:
\begin{equation}
\boldsymbol{\rho}\leftarrow
\Pi_{\|\boldsymbol{\rho}\|_\infty\leq \epsilon}
\left(\boldsymbol{\rho}+\eta
\frac{\partial \mathcal{L}_{\text{det}}(f(\mathbf{e}+\boldsymbol{\rho}),y)}
{\partial \boldsymbol{\rho}}\right),
\end{equation}
where $\Pi$ projects onto the $\ell_\infty$ ball of radius $\epsilon$ and $\eta$ is the ascent step size. The adversarial subsequence is $\mathbf{e}_{\text{adv}}=\mathbf{e}+\alpha_{\mathrm{adv}}\boldsymbol{\rho}$, with $\alpha_{\mathrm{adv}}$ scaling the perturbation before the consistency loss. By optimizing a shared perturbation direction, the regularizer discourages overly sharp decision boundaries around observed subsequences. The model is trained jointly on normal and adversarial subsequences with a composite objective:
$
\mathcal{L}_{\text{total}} =
\mathcal{L}_{\text{det}}(\mathbf{e},y)
+ \beta\, \mathcal{L}_{\text{rob}}(\mathbf{e}, \mathbf{e}_{\text{adv}}),
$
where $\mathcal{L}_{\text{rob}}(\cdot)$ is the squared difference between logits on original and adversarial inputs, and $\beta\geq0$ controls the smoothness penalty.

\begin{algorithm}[t]
\caption{Informed Reasoning Network}
\label{alg:inclair_reasoning}
\textbf{Input:} $\mathcal{D}_{\text{gold}}$, $\mathcal{D}_{\text{probe}}$, candidates $\mathcal{M}$, evaluators $\mathcal{E}_{\text{sel}}$, $\mathcal{J}$ \\
\textbf{Output:} Reasoning model $M_{\text{ctx}}$
\begin{algorithmic}[1]
\STATE \textit{// A/B testing}
\FOR{each pair $(M_a,M_b)$, $M_a\neq M_b$}
    \STATE $M_a$: few-shot on $\mathcal{D}_{\text{gold}}\rightarrow$ demos on $\mathcal{D}_{\text{probe}}$; $M_b$: demos $\rightarrow$ explain $\mathcal{D}_{\text{gold}}$; score $s_{a,b}$ via $\mathcal{E}_{\text{sel}}$
\ENDFOR
\STATE Warm-start $g(M_b)=\max_{M_a\neq M_b}s_{a,b}$; \; $Q\leftarrow\mathcal{M}$ \quad \textit{// greedy triplet ranking}
\WHILE{$|Q|>1$}
    \STATE Judges $\mathcal{J}$ score paired generators on $\mathcal{D}_{\text{gold}}$; update $g(M)$; remove $\arg\min_{M\in Q}g(M)$
\ENDWHILE
\STATE $M_{\text{best}}\leftarrow$ remaining model; generate $\mathcal{D}_{\text{syn}}$ on $\mathcal{D}_{\text{probe}}$, audit to $\mathcal{D}_{\text{syn}}^{\mathrm{audit}}$ \quad \textit{// PEFT}
\STATE Fine-tune $M_{\text{best}}$ with LoRA on $\mathcal{D}_{\text{gold}} \cup \mathcal{D}_{\text{syn}}^{\mathrm{audit}} \rightarrow M_{\text{ctx}}$; \textbf{return} $M_{\text{ctx}}$
\end{algorithmic}
\end{algorithm}

\subsection{Informed Reasoning Network}
The reasoning network uses a validation-grounded selection-and-distillation strategy rather than a single pretrained model (Algorithm~\ref{alg:inclair_reasoning}). The expert set is $\mathcal{D}_{\text{gold}} = \{(\mathbf{X}_i, y_i, \mathbf{e}_i)\}_{i=1}^{N_g}$; the reasoning stage consumes only its profile-explanation pairs $(\mathbf{X}_i, \mathbf{e}_i)$, and $\mathcal{D}_{\text{probe}} = \{\mathbf{X}_j\}$ is the unlabeled probe set.

\textbf{LLM A/B Testing.}
For each ordered pair $(M_a, M_b)$, $M_a \in \mathcal{M}$ is prompted few-shot on $\mathcal{D}_{\text{gold}}$ to generate explanations for $\mathcal{D}_{\text{probe}}$, which serve as synthetic demonstrations; $M_b \in \mathcal{M}$ is then prompted with these demonstrations to explain the annotated profiles in $\mathcal{D}_{\text{gold}}$, producing $\hat{\mathbf{e}}_i^{(a,b)} = M_b\big(\mathbf{X}_i \,\big|\, \text{few-shot from } M_a(\mathcal{D}_{\text{probe}})\big)$. The pair is scored by comparing $\hat{\mathbf{e}}_i^{(a,b)}$ with $\mathbf{e}_i$ using $\mathcal{E}_{\text{sel}}$, a selection-only clinical-rubric score for factual correctness and grounding that is disjoint from the reported automatic metrics:
\begin{equation}
s_{a,b} = \frac{1}{|\mathcal{D}_{\text{gold}}|}
\sum_{(\mathbf{X}_i,\mathbf{e}_i)\in \mathcal{D}_{\text{gold}}}
\mathcal{E}_{\text{sel}}\big(\hat{\mathbf{e}}_i^{(a,b)}, \mathbf{e}_i\big).
\end{equation}
This assesses whether a model can (i) generalize reasoning to unseen profiles and (ii) stay consistent with expert explanations when conditioned on synthetic demonstrations, filtering out brittle or non-transferable candidates. The scores warm-start the candidate ranking through $g(M_b)=\max_{M_a\neq M_b}s_{a,b}$.

\begin{table}[t]
\centering
\caption{Summary statistics of the datasets used in this study.}
\label{tab:datasets}
\resizebox{0.8\columnwidth}{!}{%
\begin{tabular}{lccc}
\toprule
\textbf{Datasets} & \textbf{\# Patients} & \textbf{\# Samples} & \textbf{Profile Length} \\
\midrule
Steroid & 29,710 & 172,260 & 3-20 \\
ADNI    & 1,308  & 2,388   & 1-8  \\
P19     & 120,000 & 1,552,211 & 4-20 \\
\bottomrule
\end{tabular}
}%
\end{table}

\begin{figure}[hbt!]
\centering{\includegraphics[width=0.9\linewidth]{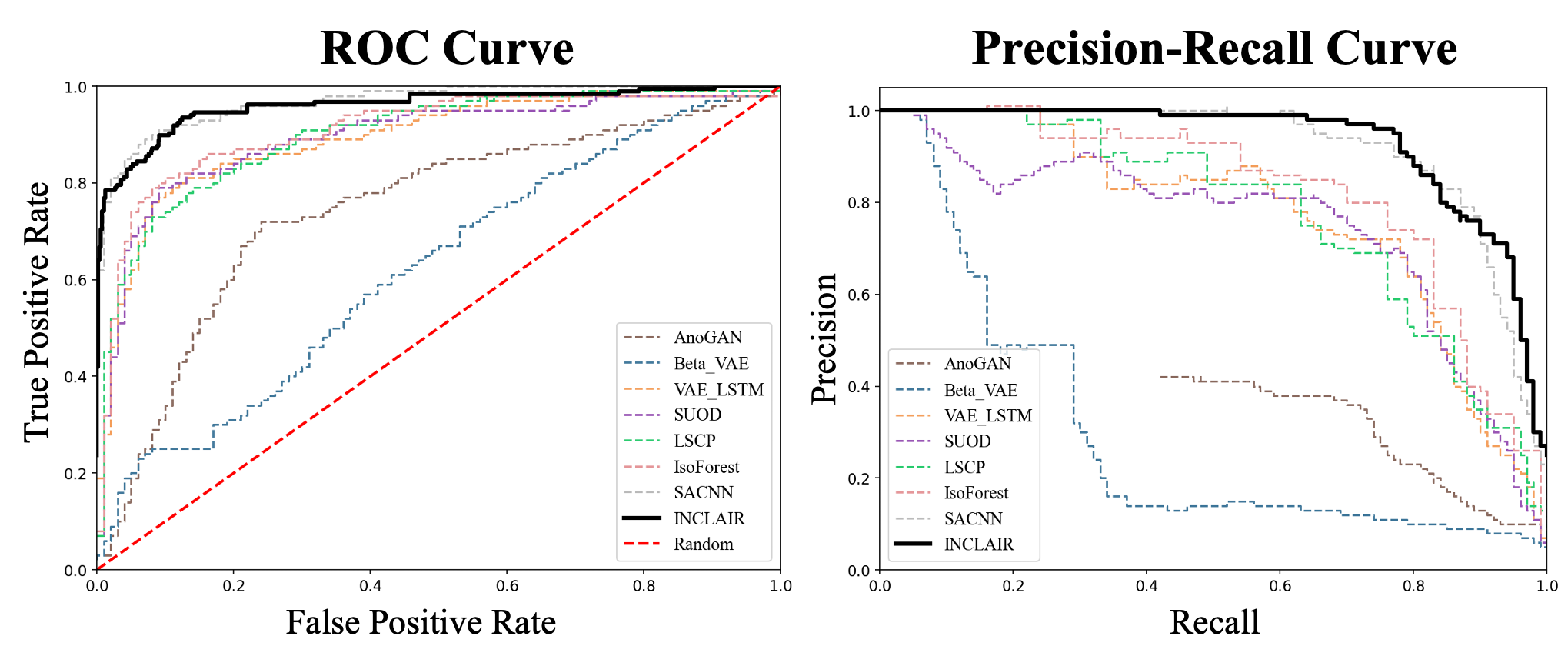}}
\caption{ROC and PR curves of INCLAIR for Steroid.
\label{roc_curve}}
\end{figure}

\textbf{Greedy Triplet Ranking.}
A/B scores can be noisy when $N_g$ is small, so INCLAIR refines the warm-started ranking of $Q$ by greedy triplet elimination. In each round, two generators $(M_u,M_v)$ explain the same gold profiles and a rotating judge $M_j$ scores factuality, temporal grounding, analyte grounding, and actionability. Each judge is calibrated against expert annotations on a held-out subset of $\mathcal{D}_{\text{gold}}$, and its scores are used only when judge-expert agreement exceeds a preset level. A candidate's score combines its A/B rubric score and its triplet win rate. This is a fixed-length elimination, not a convergence loop: the lowest-ranked $M_{\min}$ is removed each round with no re-insertion, so the procedure halts deterministically after $|\mathcal{M}|-1$ rounds. Judge rotation reduces reliance on any single evaluator but does not make agreement equivalent to correctness; $\mathcal{E}_{\text{sel}}$ therefore stays anchored to $\mathcal{D}_{\text{gold}}$, the triplet stage is evaluated as an ablation, and judge-expert agreement is reported before triplet rankings are trusted~\cite{Zheng2023MTBench}.

\textbf{Limited-Ground-Truth Analysis.}
The module's novelty is not LoRA itself but the control of selection and synthetic supervision when $N_g\ll N$. Let $\ell(M;\mathbf{X},\mathbf{e})\in[0,1]$ be a bounded expert-rubric loss with population risk $R(M)=\mathbb{E}\ell(M;\mathbf{X},\mathbf{e})$. For a finite candidate family $\mathcal{M}$, the gold-set empirical risk $\hat{R}_g(M)$ satisfies, by Hoeffding's inequality and a union bound,
$
\Pr\left(\sup_{M\in\mathcal{M}}|R(M)-\hat{R}_g(M)|>\epsilon\right)
\leq 2|\mathcal{M}|\exp(-2N_g\epsilon^2),
$
so selecting among many LLMs from few expert reports carries a complexity penalty of order $\sqrt{\log|\mathcal{M}|/N_g}$. If $\tau$ is the judge-expert disagreement rate on paired comparisons, the expected triplet ranking error is inflated by at most $\tau$ relative to an expert-only pairwise ranking, which is why we report judge-expert agreement and ablate random choice, A/B-only, and A/B-plus-triplet selection. Synthetic distillation adds a second error source. Let $\mathcal{D}_{\text{syn}}^{\mathrm{audit}}$ be the audited synthetic set and $\rho$ the probability that an accepted synthetic explanation contains an unsupported quantitative or grounding claim. With LoRA weights $w_g$ per gold rationale and $w_s\leq w_g$ per audited synthetic rationale, the synthetic supervision bias is bounded by
$
\Delta_{\text{syn}}
\leq
\frac{w_s|\mathcal{D}_{\text{syn}}^{\mathrm{audit}}|}
{w_gN_g+w_s|\mathcal{D}_{\text{syn}}^{\mathrm{audit}}|}\rho .
$
The design rule is explicit: synthetic rationales help when auditing makes $\rho$ small and gold rationales retain higher weight. Accordingly, the explanation evaluation reports numerical faithfulness, grounding agreement, the $N_g$ sweep, and the LoRA-on-gold-only ablation rather than lexical overlap alone.

\textbf{Parameter-Efficient Fine-Tuning.}
The selected model $M_{\text{best}}$ is applied few-shot to $\mathcal{D}_{\text{probe}}$, yielding a candidate synthetic set $\mathcal{D}_{\text{syn}} = \{(\mathbf{X}_j, \hat{\mathbf{e}}_j)\}_{\mathbf{X}_j \in \mathcal{D}_{\text{probe}}}$ with $\hat{\mathbf{e}}_j = M_{\text{best}}(\mathbf{X}_j)$. Each explanation is converted into structured claims (time index, analyte, direction, numeric value); claims unsupported by the structured profile, contradicting the detector attribution, or introducing unobserved clinical facts are rejected before fine-tuning. $M_{\text{best}}$ is then fine-tuned with LoRA on $\mathcal{D}_{\text{gold}} \cup \mathcal{D}_{\text{syn}}^{\mathrm{audit}}$ with audited synthetic examples down-weighted. Fine-tuning on self-generated explanations can reinforce a model's own errors, which we mitigate in three ways: synthetic rationales should pass the claim audit, they are down-weighted ($w_s \le w_g$) so their bias is bounded by $\Delta_{\text{syn}}$, and we retain a LoRA-on-gold-only control ablation.

\section{Experiments}
\subsection{Datasets}
Steroid~\cite{Rahman2022FraudEliteSports} comprises urine steroid profiles with repeated measurements of testosterone, epitestosterone, androsterone, etiocholanolone, 5$\alpha$Adiol, 5$\beta$Adiol, and their ratios, capturing individual-specific steroid-metabolism dynamics over time (Table~\ref{tab:datasets}). ADNI~\cite{Petersen2010ADNI} contains neuroimaging-derived profiles of brain structural biomarkers, including total cerebral volume, grey and white matter volumes, hippocampal volumes (left, right, total), cerebrospinal fluid volume, and white matter hyperintensity burden; these profiles are short, reflecting the sparse follow-up typical of neurodegenerative disease studies. P19~\cite{Reyna2019SepsisChallenge} is a large-scale dataset of vital-sign measurements, including heart rate, SpO$_2$, systolic and diastolic blood pressure, mean arterial pressure, and respiratory rate.

\begin{table*}[t]
\centering
\caption{Performance comparison of INCLAIR for longitudinal anomaly detection across different datasets. AC: accuracy; AUPRC: area under the precision-recall curve; SN@95SP and SN@98SP: sensitivity at 95\% and 98\% specificity; F1: F1 score; MCC: Matthews correlation coefficient; pAUC$_{0.1}$: partial AUC at a 10\% false-positive rate; AUC: area under the ROC curve. Superscripts on INCLAIR denote a statistically significant improvement over the strongest baseline in each column under a two-sided Wilcoxon signed-rank test on paired runs over identical train/test splits ($^{\ast}$: $p<0.05$; $^{\ast\ast}$: $p<0.01$).}
\label{tab:results}
\resizebox{0.8\textwidth}{!}{%
\begin{tabular}{llcccccccc}
\toprule
\textbf{Dataset} & \textbf{Method} & \textbf{AC} & \textbf{AUPRC} & \textbf{SN@95SP} & \textbf{SN@98SP} & \textbf{F1} & \textbf{MCC} & \textbf{pAUC$_{0.1}$} & \textbf{AUC} \\
\midrule
\multirow{8}{*}{Steroid}
 & Beta-VAE & 0.79$\pm$0.02 & 0.38$\pm$0.03 & 0.27$\pm$0.04 & 0.18$\pm$0.03 & 0.29$\pm$0.04 & 0.29$\pm$0.03 & 0.58$\pm$0.02 & 0.69$\pm$0.03 \\
 & V-LSTM & 0.85$\pm$0.02 & 0.52$\pm$0.03 & 0.55$\pm$0.04 & 0.42$\pm$0.04 & 0.57$\pm$0.03 & 0.53$\pm$0.03 & 0.68$\pm$0.02 & 0.79$\pm$0.02 \\
 & SUOD & 0.83$\pm$0.02 & 0.47$\pm$0.03 & 0.48$\pm$0.04 & 0.34$\pm$0.04 & 0.48$\pm$0.04 & 0.46$\pm$0.03 & 0.64$\pm$0.02 & 0.75$\pm$0.03 \\
 & LSCP & 0.82$\pm$0.02 & 0.43$\pm$0.03 & 0.44$\pm$0.04 & 0.30$\pm$0.03 & 0.44$\pm$0.03 & 0.42$\pm$0.03 & 0.61$\pm$0.02 & 0.72$\pm$0.03 \\
 & AnoGAN & 0.81$\pm$0.02 & 0.40$\pm$0.03 & 0.40$\pm$0.04 & 0.26$\pm$0.03 & 0.39$\pm$0.03 & 0.38$\pm$0.03 & 0.59$\pm$0.02 & 0.70$\pm$0.03 \\
 & IsoForest & 0.84$\pm$0.02 & 0.50$\pm$0.03 & 0.52$\pm$0.04 & 0.38$\pm$0.04 & 0.53$\pm$0.03 & 0.50$\pm$0.03 & 0.66$\pm$0.02 & 0.78$\pm$0.02 \\
 & SACNN & 0.92$\pm$0.02 & 0.81$\pm$0.02 & 0.82$\pm$0.03 & 0.75$\pm$0.03 & 0.83$\pm$0.02 & 0.79$\pm$0.03 & 0.86$\pm$0.02 & 0.92$\pm$0.02 \\
 & \textbf{INCLAIR} & \textbf{0.95$\pm$0.01}$^{\ast}$ & \textbf{0.88$\pm$0.02}$^{\ast\ast}$ & \textbf{0.90$\pm$0.02}$^{\ast\ast}$ & \textbf{0.84$\pm$0.03}$^{\ast\ast}$ & \textbf{0.88$\pm$0.02}$^{\ast\ast}$ & \textbf{0.85$\pm$0.02}$^{\ast\ast}$ & \textbf{0.91$\pm$0.01}$^{\ast\ast}$ & \textbf{0.96$\pm$0.01}$^{\ast\ast}$ \\
\midrule
\multirow{8}{*}{ADNI}
 & Beta-VAE & 0.55$\pm$0.02 & 0.52$\pm$0.04 & 0.15$\pm$0.03 & 0.07$\pm$0.02 & 0.13$\pm$0.03 & 0.12$\pm$0.03 & 0.54$\pm$0.03 & 0.57$\pm$0.04 \\
 & V-LSTM & 0.58$\pm$0.03 & 0.58$\pm$0.04 & 0.25$\pm$0.04 & 0.12$\pm$0.03 & 0.21$\pm$0.04 & 0.20$\pm$0.03 & 0.61$\pm$0.03 & 0.66$\pm$0.04 \\
 & SUOD & 0.57$\pm$0.02 & 0.55$\pm$0.04 & 0.22$\pm$0.04 & 0.10$\pm$0.03 & 0.18$\pm$0.03 & 0.17$\pm$0.03 & 0.58$\pm$0.03 & 0.61$\pm$0.04 \\
 & LSCP & 0.56$\pm$0.02 & 0.53$\pm$0.04 & 0.20$\pm$0.04 & 0.09$\pm$0.02 & 0.16$\pm$0.03 & 0.16$\pm$0.03 & 0.56$\pm$0.03 & 0.59$\pm$0.04 \\
 & AnoGAN & 0.56$\pm$0.02 & 0.52$\pm$0.04 & 0.18$\pm$0.04 & 0.08$\pm$0.02 & 0.15$\pm$0.03 & 0.14$\pm$0.03 & 0.55$\pm$0.03 & 0.58$\pm$0.04 \\
 & IsoForest & 0.57$\pm$0.02 & 0.56$\pm$0.04 & 0.23$\pm$0.04 & 0.11$\pm$0.03 & 0.19$\pm$0.03 & 0.19$\pm$0.03 & 0.59$\pm$0.03 & 0.63$\pm$0.04 \\
 & SACNN & 0.57$\pm$0.03 & 0.55$\pm$0.03 & 0.21$\pm$0.04 & 0.10$\pm$0.03 & 0.18$\pm$0.03 & 0.17$\pm$0.03 & 0.57$\pm$0.03 & 0.60$\pm$0.04 \\
 & \textbf{INCLAIR} & \textbf{0.63$\pm$0.03}$^{\ast}$ & \textbf{0.68$\pm$0.03}$^{\ast\ast}$ & \textbf{0.38$\pm$0.04}$^{\ast\ast}$ & \textbf{0.24$\pm$0.04}$^{\ast\ast}$ & \textbf{0.38$\pm$0.04}$^{\ast\ast}$ & \textbf{0.33$\pm$0.04}$^{\ast\ast}$ & \textbf{0.70$\pm$0.03}$^{\ast\ast}$ & \textbf{0.76$\pm$0.03}$^{\ast\ast}$ \\
\midrule
\multirow{8}{*}{P19}
 & Beta-VAE & 0.91$\pm$0.01 & 0.10$\pm$0.02 & 0.11$\pm$0.02 & 0.06$\pm$0.02 & 0.09$\pm$0.02 & 0.07$\pm$0.02 & 0.52$\pm$0.02 & 0.56$\pm$0.03 \\
 & V-LSTM & 0.92$\pm$0.01 & 0.13$\pm$0.02 & 0.18$\pm$0.03 & 0.10$\pm$0.02 & 0.15$\pm$0.02 & 0.13$\pm$0.02 & 0.57$\pm$0.02 & 0.64$\pm$0.03 \\
 & SUOD & 0.92$\pm$0.01 & 0.12$\pm$0.02 & 0.16$\pm$0.03 & 0.08$\pm$0.02 & 0.12$\pm$0.02 & 0.10$\pm$0.02 & 0.55$\pm$0.02 & 0.61$\pm$0.03 \\
 & LSCP & 0.92$\pm$0.01 & 0.11$\pm$0.02 & 0.14$\pm$0.03 & 0.07$\pm$0.02 & 0.11$\pm$0.02 & 0.08$\pm$0.02 & 0.54$\pm$0.02 & 0.59$\pm$0.03 \\
 & AnoGAN & 0.91$\pm$0.01 & 0.09$\pm$0.02 & 0.10$\pm$0.02 & 0.05$\pm$0.02 & 0.08$\pm$0.02 & 0.05$\pm$0.02 & 0.51$\pm$0.02 & 0.55$\pm$0.03 \\
 & IsoForest & 0.92$\pm$0.01 & 0.12$\pm$0.02 & 0.17$\pm$0.03 & 0.09$\pm$0.02 & 0.13$\pm$0.02 & 0.12$\pm$0.02 & 0.56$\pm$0.02 & 0.62$\pm$0.03 \\
 & SACNN & 0.92$\pm$0.01 & 0.19$\pm$0.02 & 0.28$\pm$0.03 & 0.18$\pm$0.03 & 0.25$\pm$0.03 & 0.24$\pm$0.03 & 0.63$\pm$0.02 & 0.70$\pm$0.03 \\
 & \textbf{INCLAIR} & \textbf{0.93$\pm$0.01} & \textbf{0.31$\pm$0.02}$^{\ast\ast}$ & \textbf{0.49$\pm$0.04}$^{\ast\ast}$ & \textbf{0.34$\pm$0.03}$^{\ast\ast}$ & \textbf{0.42$\pm$0.03}$^{\ast\ast}$ & \textbf{0.41$\pm$0.03}$^{\ast\ast}$ & \textbf{0.73$\pm$0.02}$^{\ast\ast}$ & \textbf{0.78$\pm$0.03}$^{\ast\ast}$ \\
\bottomrule
\end{tabular}
}%
\end{table*}

\begin{figure*}[hbt!]
\centering{\includegraphics[width=0.7\textwidth]{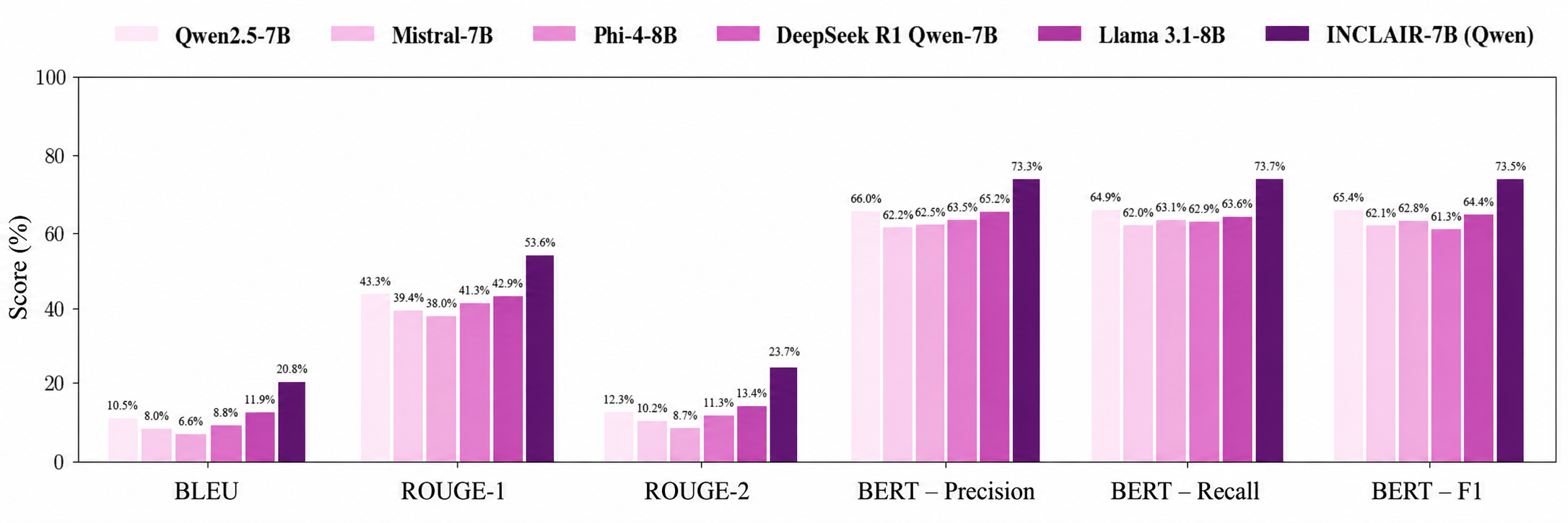}}
\caption{Performance comparison of INCLAIR and general-purpose LLMs  for informed reasoning.
\label{results_informed_reasoning}}
\end{figure*}

\subsection{Baseline Methods}
\textbf{Anomaly Detection.}
We compare INCLAIR with Beta-VAE~\cite{Higgins2017BetaVAE} and AnoGAN~\cite{Schlegl2017GANAnomalyDetection} as generative reconstruction-based methods, V-LSTM~\cite{Malhotra2016LSTMPrognostics} as a temporal forecasting model, SUOD~\cite{Zhao2021SUOD} and LSCP~\cite{Zhao2019LSCP} as unsupervised ensemble detectors, IsoForest~\cite{Liu2012IsolationForest} as a classical baseline, and SACNN~\cite{Rahman2024SACNN} as a strong deep learning model for subsequence-based longitudinal anomaly detection.

\textbf{Informed Reasoning.}
We compare INCLAIR against domain-specific biomedical LLMs and strong general-purpose models. Biomedical baselines include BioMistral-7B~\cite{Labrak2024BioMistral}, ClinicalGPT-7B~\cite{Wang2023ClinicalGPT}, AlpaCare-8B~\cite{Zhang2023AlpacaRE}, OpenBioLLM-8B~\cite{OpenBioLLMs}, MedAlpaca-7B~\cite{Han2023MedAlpaca}, and Meditron-7B~\cite{Chen2023Meditron70B}. General-purpose LLMs include Qwen2.5-7B~\cite{Hui2024Qwen25Coder}, Mistral-7B~\cite{Jiang2024Mixtral}, Phi-4-8B~\cite{Abdin2024Phi4}, DeepSeek R1-7B~\cite{Guo2025DeepSeekR1}, and Llama 3.1-8B~\cite{Grattafiori2024LLaMA3}.

\subsection{Experimental Settings}
We use a patient-level hold-out protocol: profiles, not individual observations, are split into disjoint training, validation, and test sets so no patient appears in more than one split, with results reported as mean and standard deviation over five random splits. Every hyperparameter is selected on validation; test labels are used only for final evaluation. We set $l=5$, project each $l \times (p+2)$ subsequence to 16 channels, and process it with stacked inception blocks of increasing kernel size; $F_{\text{thres}}$ targets 95\% specificity (SN@95SP). As the maximum profile length is 20, full combinatorial enumeration is tractable and used by default, with the Corollary~1 sampler as the scalable alternative for longer profiles. For the reasoning pipeline, the candidate family $\mathcal{M}$ is the biomedical and general-purpose 7-8B LLMs above; the selection procedure returns Qwen2.5-7B as $M_{\text{best}}$, LoRA fine-tuned to produce INCLAIR-7B. All selection signals ($\mathcal{E}_{\text{sel}}$, the A/B scores $s_{a,b}$, and the LLM-as-judge $\mathcal{J}$) are computed only on $\mathcal{D}_{\text{gold}}$ and are disjoint from $\mathcal{D}_{\text{syn}}$; the metrics in Table~\ref{tab:reasoning_results} use a held-out split of $\mathcal{D}_{\text{gold}}$ not used for selection or fine-tuning.

\begin{table*}[t]
\centering
\caption{Performance comparison of INCLAIR and biomedical LLM baselines for informed reasoning. Superscripts on INCLAIR denote a statistically significant improvement over the strongest baseline in each column under a two-sided Wilcoxon signed-rank test on paired runs over identical evaluation splits ($^{\ast}$: $p<0.05$; $^{\ast\ast}$: $p<0.01$).}
\label{tab:reasoning_results}
\resizebox{0.7\textwidth}{!}{%
\begin{tabular}{lcccccc}
\toprule
\textbf{Baselines} & \textbf{BLEU(\%)} & \multicolumn{2}{c}{\textbf{ROUGE(\%)}} & \multicolumn{3}{c}{\textbf{BERTScore(\%)}} \\
\cmidrule(lr){3-4} \cmidrule(lr){5-7}
 &  & ROUGE-1 & ROUGE-2 & Precision & Recall & F1 \\
\midrule
BioMistral-7B   & 5.42$\pm$0.63  & 33.71$\pm$1.24 & 7.43$\pm$0.82  & 58.92$\pm$1.53 & 54.41$\pm$1.62 & 56.53$\pm$1.44 \\
ClinicalGPT-7B & 4.91$\pm$0.54  & 32.53$\pm$1.17 & 8.63$\pm$0.94  & 60.91$\pm$1.46 & 53.62$\pm$1.57 & 56.93$\pm$1.34 \\
AlpaCare-8B    & 9.24$\pm$0.73  & 40.32$\pm$1.36 & 11.53$\pm$1.04 & 64.93$\pm$1.27 & 61.94$\pm$1.31 & 63.42$\pm$1.16 \\
OpenBioLLM-8B  & 5.56$\pm$0.61  & 35.84$\pm$1.25 & 12.82$\pm$1.14 & 68.54$\pm$1.18 & 58.47$\pm$1.42 & 63.02$\pm$1.28 \\
MedAlpaca-7B   & 9.27$\pm$0.83  & 38.56$\pm$1.47 & 10.54$\pm$0.96 & 63.37$\pm$1.36 & 61.13$\pm$1.24 & 62.18$\pm$1.14 \\
Meditron-7B    & 6.94$\pm$0.67  & 41.05$\pm$1.35 & 10.96$\pm$1.08 & 60.62$\pm$1.58 & 62.34$\pm$1.37 & 61.46$\pm$1.26 \\
\midrule
\textbf{INCLAIR-7B} & \textbf{20.84$\pm$0.93}$^{\ast\ast}$ & \textbf{53.64$\pm$1.57}$^{\ast\ast}$ & \textbf{23.74$\pm$1.26}$^{\ast\ast}$ & \textbf{73.31$\pm$1.14}$^{\ast\ast}$ & \textbf{73.72$\pm$1.07}$^{\ast\ast}$ & \textbf{73.54$\pm$1.06}$^{\ast\ast}$ \\
\bottomrule
\end{tabular}
}%
\end{table*}

\begin{table*}[t]
\centering
\caption{Ablation study results on the Steroid dataset.}
\label{tab:ablation_results}
\resizebox{0.8\textwidth}{!}{%
\begin{tabular}{llcccccc}
\toprule
\textbf{Ablation} &
\textbf{Variant} &
\textbf{AUPRC} &
\textbf{SN@98SP} &
\textbf{F1} &
\textbf{MCC} &
\textbf{pAUC$_{0.1}$} &
\textbf{AUC} \\
\midrule

\multirow{2}{*}{Architecture}
& \textit{w/o} Inception block
& 0.74$\pm$0.03 & 0.66$\pm$0.04 & 0.77$\pm$0.03
& 0.73$\pm$0.03 & 0.78$\pm$0.02 & 0.88$\pm$0.02 \\
& \textit{w} Inception block
& 0.81$\pm$0.02 & 0.75$\pm$0.03 & 0.83$\pm$0.02
& 0.79$\pm$0.03 & 0.86$\pm$0.02 & 0.92$\pm$0.02 \\
\midrule

\multirow{3}{*}{History construction}
& Sliding window
& 0.77$\pm$0.03 & 0.69$\pm$0.04 & 0.79$\pm$0.03
& 0.75$\pm$0.03 & 0.81$\pm$0.02 & 0.89$\pm$0.02 \\
& Random-$B$ subsets
& 0.86$\pm$0.02 & 0.82$\pm$0.03 & 0.87$\pm$0.02
& 0.84$\pm$0.02 & 0.90$\pm$0.01 & 0.95$\pm$0.01 \\
& Full combinatorial
& 0.88$\pm$0.02 & 0.84$\pm$0.03 & 0.88$\pm$0.02
& 0.85$\pm$0.02 & 0.91$\pm$0.01 & 0.96$\pm$0.01 \\
\midrule

\multirow{3}{*}{Aggregation}
& Mean pooling
& 0.80$\pm$0.03 & 0.70$\pm$0.04 & 0.80$\pm$0.03
& 0.76$\pm$0.03 & 0.83$\pm$0.02 & 0.91$\pm$0.02 \\
& Max pooling
& 0.84$\pm$0.02 & 0.80$\pm$0.03 & 0.85$\pm$0.02
& 0.82$\pm$0.02 & 0.88$\pm$0.02 & 0.94$\pm$0.01 \\
& Top-$k$ pooling
& 0.88$\pm$0.02 & 0.84$\pm$0.03 & 0.88$\pm$0.02
& 0.85$\pm$0.02 & 0.91$\pm$0.01 & 0.96$\pm$0.01 \\
\midrule

\multirow{2}{*}{Smoothness}
& \textit{w/o} Regularization
& 0.85$\pm$0.02 & 0.80$\pm$0.03 & 0.86$\pm$0.02
& 0.82$\pm$0.02 & 0.88$\pm$0.02 & 0.94$\pm$0.01 \\
& \textit{w} Regularization
& 0.88$\pm$0.02 & 0.84$\pm$0.03 & 0.88$\pm$0.02
& 0.85$\pm$0.02 & 0.91$\pm$0.01 & 0.96$\pm$0.01 \\
\midrule

\multirow{3}{*}{Window length}
& $l=3$
& 0.84$\pm$0.02 & 0.79$\pm$0.03 & 0.85$\pm$0.02
& 0.81$\pm$0.02 & 0.88$\pm$0.02 & 0.94$\pm$0.01 \\
& $l=5$
& 0.88$\pm$0.02 & 0.84$\pm$0.03 & 0.88$\pm$0.02
& 0.85$\pm$0.02 & 0.91$\pm$0.01 & 0.96$\pm$0.01 \\
& $l=8$
& 0.82$\pm$0.03 & 0.76$\pm$0.03 & 0.83$\pm$0.03
& 0.79$\pm$0.03 & 0.86$\pm$0.02 & 0.93$\pm$0.02 \\

\bottomrule
\end{tabular}%
}
\end{table*}

\begin{figure*}[hbt!]
\centering{\includegraphics[width=0.8\textwidth]{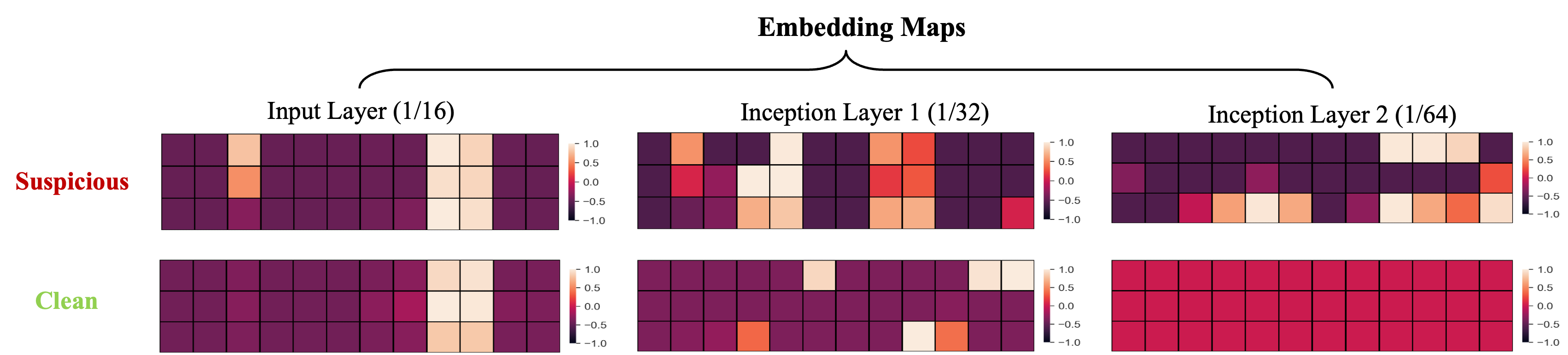}}
\caption{Embedding maps of the input and successive inception layers for a suspicious and a clean profile.
\label{case_study:embeddings}}
\end{figure*}

\section{Results}
\subsection{Anomaly Detection}
As shown in Table~\ref{tab:results}, INCLAIR outperforms all baselines on every metric across the three datasets. On Steroid it attains the highest accuracy (0.95) and AUC (0.96) with a substantial sensitivity gain at high specificity; the trend holds on ADNI (AUC 0.76) despite sparse, short profiles and on large-scale P19 (accuracy 0.93, AUC 0.74), where it clearly exceeds classical and deep baselines in sensitivity. Its high-specificity sensitivity on ADNI remains modest (SN@95SP $=0.38$), which we attribute to the short ADNI profiles (length 1 to 8; Table~\ref{tab:datasets}) that yield few historical contexts and, by Proposition~1, inflate profile-score variance; this regime is therefore best used as a sensitivity-oriented triage aid with further follow-up rather than a standalone screening decision. The Steroid ROC and PR curves (Figure~\ref{roc_curve}) confirm this: INCLAIR outperform every baseline, rising steeply at low false-positive rates and staying above SACNN throughout, and on the PR curve sustaining near-perfect precision at low-to-mid recall while degrading more slowly than all baselines toward high recall; the generative models (Beta-VAE, AnoGAN) collapse early and the ensembles (SUOD, LSCP) and IsoForest trail across the operating range.

\subsection{Informed Reasoning}
The reasoning evaluation is restricted to the Steroid dataset, the only cohort with expert-written natural-language rationales and DNA-based confirmatory labels. Table~\ref{tab:reasoning_results} compares explanation quality against biomedical LLMs: INCLAIR attains the highest BERTScore, indicating explanations that are more accurate and complete with respect to expert annotations, whereas the biomedical baselines show moderate lexical overlap but limited semantic consistency. Figure~\ref{results_informed_reasoning} extends the comparison to strong general-purpose LLMs, over which INCLAIR again leads on both lexical overlap and semantic similarity; models such as Qwen2.5-7B, Mistral-7B, and Phi-4-8B reach only moderate BERTScore, reflecting weaker contextual coherence. These results show that INCLAIR leverages limited expert supervision to produce explanations better aligned with domain-expert reasoning.

\subsection{Ablation Studies}
Table~\ref{tab:ablation_results} reports ablations on the Steroid dataset, and each design choice contributes as the theory predicts. The inception block raises AUPRC (0.74-0.81) and AUC (0.88-0.92), confirming the value of multi-scale temporal features. For history construction, full combinatorial enumeration is best (AUPRC 0.88, AUC 0.96) and the random-$B$ subset sampler of Corollary~1 nearly matches it (0.86, 0.95) while both clearly exceed a single sliding window (0.77, 0.89), validating the incomplete-subset approximation. For aggregation, top-$k$ pooling improves high-specificity sensitivity over mean pooling (SN@98SP 0.84 vs.\ 0.70) and over max pooling (0.80), consistent with Proposition~2. Adversarial smoothness regularization adds a further gain (AUPRC 0.85-0.88), and the window length is optimal at $l=5$, above $l=3$, matching the variance-window trade-off of Proposition~1.

\section{Case Study}
We evaluated 29 longitudinal steroid profiles against DNA ground truth: 7 anomalous (2 sample swaps, 5 doping) and 22 clean. INCLAIR recovered both sample swaps and all 22 clean profiles, matching the DNA conclusions, and detected 3/5 doping cases, reflecting the difficulty of separating subtle metabolic manipulation from physiological variability. Given the small cohort we report exact outcomes rather than rates: these are illustrative field results, above the chance level implied by the anomaly base rate but too few for confidence intervals, so the quantitative comparisons in Table~\ref{tab:results} remain the primary evidence. The embedding maps (Figure~\ref{case_study:embeddings}) support this qualitatively: the suspicious and clean profiles are comparably variable at the input layer, but through the inception layers the suspicious profile develops sparse, localized activations indicating abrupt deviations while the clean profile stays smooth and evenly distributed.

\section{Conclusion}
We presented INCLAIR, a structure-aware framework for anomaly detection and informed reasoning in longitudinal clinical profiles. Casting the profile score as an order-$l$ U-statistic yields a variance decomposition, a sparse-anomaly attenuation result motivating top-$k$ pooling, and an incomplete-subset approximation that bounds inference cost. Extending the reasoning module to cohorts without expert rationales and adding population-referenced normalization for chronic deviations are natural future directions.

\pagebreak
\bibliography{aaai2027}


\end{document}